\documentclass[runningheads]{llncs}

\usepackage{eccv}

\usepackage{eccvabbrv}

\usepackage{graphicx}
\usepackage{booktabs}

\usepackage[accsupp]{axessibility}  % Improves PDF readability for those with disabilities.

\usepackage[pagebackref,breaklinks,colorlinks,citecolor=eccvblue]{hyperref}
\usepackage{orcidlink}

\usepackage[dvipsnames]{xcolor}
\usepackage{times}
\usepackage{epsfig}
\usepackage{graphicx}
\usepackage{amsmath}
\usepackage{amssymb}
\usepackage{booktabs}
\usepackage{adjustbox}
\usepackage{multirow}
\usepackage{pifont}
\usepackage{comment}
\usepackage{subcaption}
\usepackage{arydshln} % for dashed lines
\usepackage{appendix}
\usepackage{wrapfig}

\newcommand{\datasetsize}{\textbf{50,000}}

\newcommand{\taskname}{{Affective Visual Dialog}~}
\newcommand{\datasetshortname}{\textbf{AffectVisDial}}
\newcommand{\datasetshortnamenb}{{AffectVisDial}~}

\newcommand{\dialognum}{50K~} %30803
\newcommand{\qnum}{500K~}
\newcommand{\qapair}{500,000~}
\newcommand{\alen}{14.85~} %9.82856
\newcommand{\qlen}{10.92~} %8.323
\newcommand{\uniquea}{324K~} %146645
\newcommand{\uniqueq}{193K~} %45195
\newcommand{\answertopcover}{38\%~} %0.37698
\newcommand{\explenbefore}{27.29~} %30.910
\newcommand{\explenafter}{28.95~} %25.988
\newcommand{\vocab}{21k~} %21233

\begin{document}

\title{Affective Visual Dialog: A Large-Scale Benchmark for Emotional Reasoning Based on Visually Grounded Conversations} 

\titlerunning{Affective Visual Dialog}


\author{Kilichbek Haydarov\inst{1} \and
Xiaoqian Shen\inst{1} \and
Avinash Madasu\inst{1} \and Mahmoud Salem\inst{1} \and Li-Jia Li\inst{2} \and Gamaleldin Elsayed\inst{3}
\and Mohamed Elhoseiny\inst{1}}

\authorrunning{K.~Haydarov et al.}

\institute{King Abdullah University of Science and Technology \and HealthUnity \and Google DeepMind}

\maketitle

\begin{abstract}
  We introduce Affective Visual Dialog, an emotion explanation and reasoning task as a testbed for research on understanding constructed emotions in response to visually grounded conversations. The task involves three skills: (1) Dialog-based Question Answering (2) Dialog-based Emotion Prediction and (3) Affective explanation generation based on the dialog.  Our key contribution is the collection of a large-scale dataset, dubbed AffectVisDial, consisting of 50K 10-turn visually grounded dialogs as well as concluding emotion attributions and dialog-informed textual emotion explanations, resulting in a total of 27,180 working hours. Notably, the dataset spans a broad range of visual stimuli, covering human heritage and contemporary life, with an average per-turn answer length of about 12 words — 5 times that of the VisDial dataset — and explanations exceeding 28 words on average. We explain our determining design decisions in collecting the dataset, data inclusion and exclusion criteria starting from over  100K dialogs for quality control,  and introduce the questioner and answerer tasks that are associated with the participants in the conversation.  We propose and demonstrate solid \taskname{}baselines adapted from state-of-the-art multimodal models. Remarkably, the responses generated by our models show promising emotional reasoning abilities in response to visually grounded conversations. Our project page with the dataset is available through \url{https://affective-visual-dialog.github.io}

\end{abstract}

\section{Introduction}
\label{sec:intro}

As AI systems become increasingly ubiquitous, it is crucial to consider the emotional aspects of human nature to develop systems that can flexibly respond naturally based on perceived emotions, ultimately increasing the social acceptance of AI and better-supporting humans. Stuart Russel~\cite{russell2019human} critiqued the normative AI development model warning of potential catastrophes due to its frequent disregard of compatibility with human values and goals. Developing emotion-aware AI systems requires public datasets capturing diverse sensory modalities. In this work, our focus is to take a step towards developing Computer Vision systems compatible with our emotional being.

In recent years, there has been growing attention to examining the emotional impact of visual stimuli on viewers, expressed via language~\cite{achlioptas2021artemis, mohamed2022okay, mohamed2022artelingo, Achlioptas_2023_CVPR}. While these works mostly focused on the direct influence of visual stimuli on affective experiences, there is a notable gap in understanding how \textit{language grounded in visuals} influences affective experiences. Our research aims to bridge this gap by investigating the construction of emotions in visually grounded conversations and elucidating how such emotions are conveyed and explained through language. 

\begin{wrapfigure}{r}{0.5\textwidth}
\begin{center}
  \vspace{-15mm}
   \includegraphics[width=1.0\linewidth]{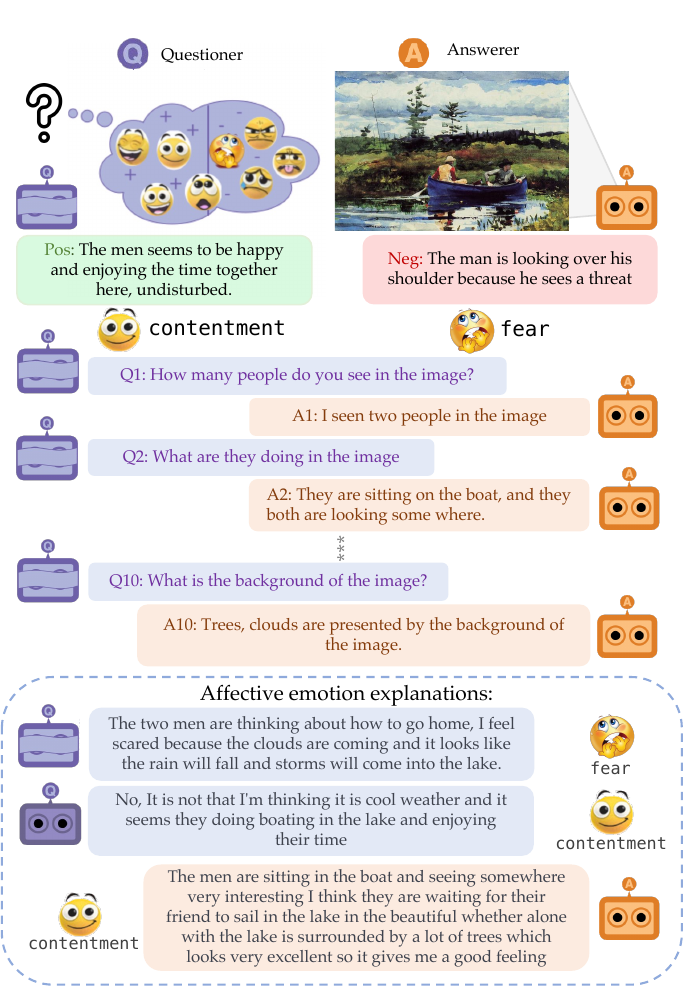}
  %\vspace{-2mm}
   \caption{\datasetshortname{} captures constructed emotion attributions and explanations from both the Questioner (without image access) and Answerer (with image access) after 10 turns of questions and answers starting from two opposing opinions.
   Subsequently, the Questioner views the image and can alter their initial emotional response, accompanied by a corresponding textual explanation.
} \label{figure:teaser}
\end{center}
\vspace{-5mm}
\end{wrapfigure}

Our effort starts with building a large-scale dataset that captures language and vision modalities, and their influence on human emotions. Specifically, we are interested in language in the form of a dialog about a visual signal that guides human emotion. The dialogues capture live conversations between two people and their constructed emotions in response to visual stimuli.
Datasets of this nature aid in deepening our understanding of the correlation between vision (and language) information and emotion. It serves as a valuable resource for refining models and their evaluation. This dataset not only contributes to our comprehension of the link between visual and linguistic cues in emotion recognition but also empowers us to construct more adept AI models with enhanced emotion understanding and evaluate their performance more effectively.
To that end, we design a novel task to establish a dialogue between two human participants - a \emph{Questioner} and an \emph{Answerer} - to reach an emotional judgment. Our task allows models to generate emotions and explanations through visually grounded conversations and the integration of vision and language modalities.
Our dataset provides an avenue to explore the impact of direct/indirect (e.g. image is not visible) access to visual stimuli on affective experiences. We can explore emotional states after the conversation about the hidden image and how they may change once the image is revealed. For example, Fig.~\ref{figure:teaser} shows a shift in the questioner's emotional response from \textit{``fear"} to \textit{``contentment"} once they saw the hidden visual stimuli. 
Furthermore, our dataset enables us to look at the impact of visual stimuli on the Answerer's emotional state after the conversation.

\textbf{Why Dialog setup?} The question-answer-based dialog setup was chosen over providing text descriptions about visual stimuli because this setup offers unique advantages that enable a more dynamic and interactive exploration of the visual content. This allows the questioner to obtain specific information, seek clarification, and delve deeper into particular aspects of the content, leading to a more comprehensive understanding of the visual stimulus. This in turn improves certain abilities of AI agents supported
by research~\cite{murahari-etal-2019-improving, zhu2023chatgpt, chen2023video}.

We present a benchmark, \datasetshortname{}, with standard splits and baseline models for the \emph {Affective Visual Dialog} Task: evaluating (1) Question Answering ability and (2) Emotion Classification based on visually grounded dialogs (3) and Affective Explanation Generation from dialogs.
\noindent In summary, our contributions include:
\begin{itemize}
   \item We introduce a large-scale dialog dataset  (\datasetshortname{}) capturing conversations between two participants along with their emotional reactions, encompassing ~\datasetsize{} unique dialogs with emotion judgments and explanations.
   
   \item We design a novel task of \taskname{} covering three subtasks: 1) Dialog-based question answering, 2) Dialog-based emotion classification, and (3) Affective Explanation where the system needs to produce emotion and explanation in language based on dialogs and visuals.

   \item We introduce several baselines and measure their ability to predict emotions based on language and/or vision, and most importantly explain \textit{why}.

   \item Our findings show that emotion can be influenced by the different responses within the dialogue, demonstrating the importance of capturing the conversation context based on visuals for emotion prediction. 
\end{itemize}
We conclude by illustrating in the discussion how the introduced models could potentially guide conversations according to a desired emotion, and edit the visual stimuli based on the stipulated emotional reasoning.

\section{Related Work}
% Vision-Language
In this section, we survey prior research in vision language datasets, affective explanation, and dialog systems, emphasizing the opportunities that our dataset offers to advance these areas.

\noindent \textbf{Vision-Language datasets} (e.g.~\cite{lin2014microsoft, antol2015vqa, sharma2018conceptual, young2014image, das2017visual, Vries_2017_CVPR}) have been curated to explore the link between visual and linguistic information, facilitating the investigation of tasks such as image captioning, visual question answering (VQA), and visual dialog.
The latter two tasks challenge machines to reason over visual elements and general knowledge to infer correct answers based on images and related questions. These demanding tasks require machines to exhibit visual understanding, reasoning, and text comprehension. Our work extends these challenges by incorporating affective reasoning and explanation generation capabilities.

\noindent \textbf{Emotion Representation \& Dialog Datasets.} In the existing literature, two widely adopted paradigms for representing emotions are the discrete categorical system~\cite{ekman1992argument} and the continuous 2D-dimensional Valence-Arousal (VA) model~\cite{russell1980circumplex}. Drawing from prior works (e.g.,\cite{you2016building, yanulevskaya2008emotional, machajdik2010affective, achlioptas2021artemis, mohamed2022okay, mohamed2022artelingo}), we follow a consistent methodology and leverage the same set of eight Ekman emotion categories: the four negative emotions (anger, disgust, fear, sadness) are considered universal and basic, as initially proposed by Ekman, the four positive emotions (amusement, awe, contentment, excitement) are more nuanced variations of happiness~\cite{diener2009evolving}. Numerous efforts have been made to collect datasets and analyze human emotions across different modalities, encompassing facial expression~\cite{mollahosseini2017affectnet}, body gestures~\cite{liu2021imigue, gunes2006bimodal, ranganathan2016multimodal, goyal2017something}, text~\cite{strapparava2007semeval,demszky2020goemotions, buechel2022emobank}, music~\cite{bogdanov2019mediaeval, fan2017emo, hung2021emopia, cowen2020music} and visual art~\cite{achlioptas2021artemis, mohamed2022okay}. However, most of them were limited in size or concerned with single modalities. Although some recent studies have explored the link between emotion attributes, language, and visual stimuli~\cite{achlioptas2021artemis, Achlioptas_2023_CVPR, mohamed2022artelingo}, they often lack explanatory language based on linguistic cues from conversations surrounding hidden visual stimuli. While there are dialog datasets that address human emotions and social intelligence~\cite{li2017dailydialog, chen2018emotionlines, zhou2023sotopia, sap2019social, li2024diplomat, rashkin2019towards}, they lack grounding in visual signals, which sets our work apart. Our dataset aims to bridge this gap and is, to the best of our knowledge, the first to focus on capturing emotional responses and affective expressions through dialogues centered around visual stimuli.

\section{\datasetshortname{} Dataset and Analysis}

\subsection{Dataset Collection} 
\begin{wrapfigure}{l}{0.45\textwidth}
    \begin{center}
    \vspace{-4mm}
    \includegraphics[width=0.45\textwidth]{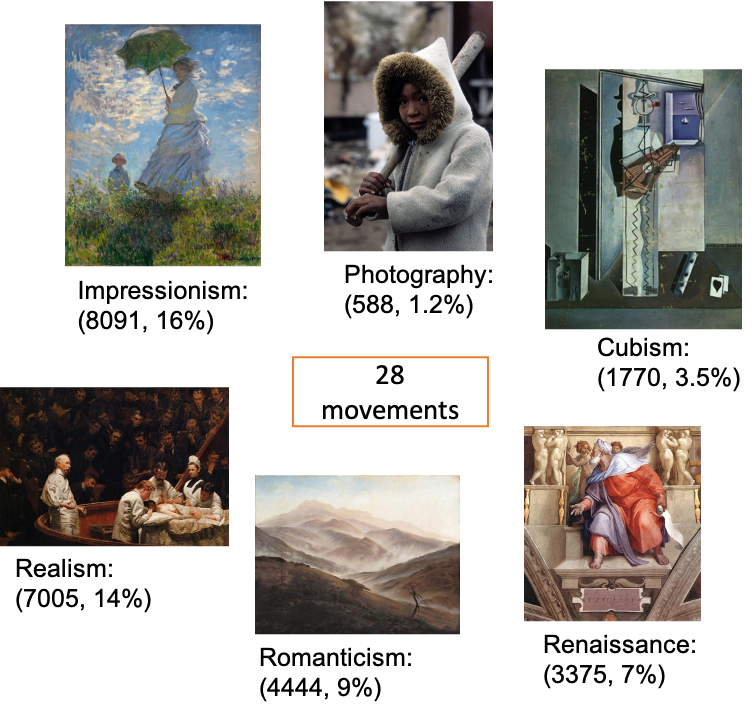}
   \caption{Visual stimuli from diverse movements including photography; dialog counts and  percentages are in parentheses.} \label{figure:visual_stimuli}
    \smallskip\par
     \smallskip\par
   \includegraphics[width=0.45\textwidth]{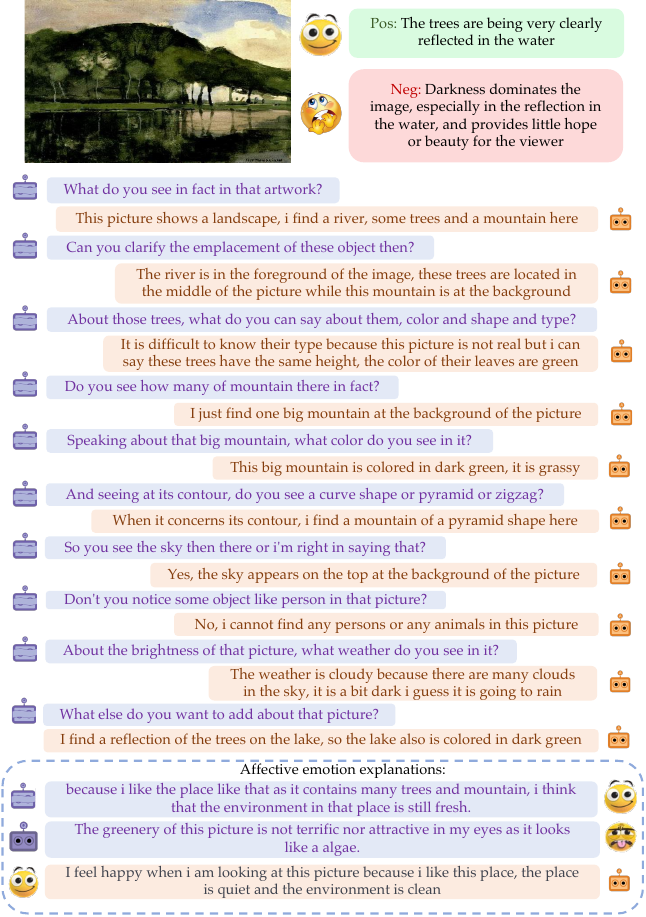}
   \caption{An example sample from \datasetshortnamenb{} dataset. It contains an image, two opinions (positive and negative) about the image, a conversation, and explanations from both Questioner and Answerer.}
\vspace{-25mm}
   \label{figure:example_dialogs}
    \end{center}
\end{wrapfigure}

We employed a live communication protocol, where two agents engage in a dialogue. 
The Questioner asks questions about a hidden image, which is intentionally concealed to mitigate any visual priming biases. These biases can cause models to operate under the assumption that the questioner will inquire about the objects depicted in the image~\cite{agrawal2018don, goyal2017making}. The objective of the Questioner is to explore the hidden image. On the other hand, the second agent (Answerer) observes the image and provides answers to the questions posed by the Questioner. The unique setup in our user interface design is that to initiate the conversation, we reveal two opposing opinions (a negative and a positive caption) from the combined ArtEmis v1 and v2 datasets~\cite{achlioptas2021artemis, mohamed2022okay} associated with the artwork. Our intuition of triggering the dialog with the two opposite opinions is to counter the biases of the Questioner when starting the task and encourage more open-mindedness towards the emotion triggered by the hidden visual signal (i.e., to open the possibility that the hidden visuals may be perceived subjectively positively or negatively). After 10 question-answer pair exchanges, the Questioner is asked to provide an emotional class response and explanation for what informed the Questioner's decision based on the dialog. Then, the hidden image is revealed to the Questioner, and both agents are required to indicate their emotional responses in the presence of visual cues and conversation. This final question after revealing the image allows explorations regarding the emotion arising from only dialog vs. those that are also informed by visual stimuli. For the live chat framework, we use Mephisto tool~\cite{mephisto} with our customized front-end interfaces. Fig.~\ref{figure:example_dialogs} shows an example of collected dialog; more examples and interfaces are attached in the supplementary. 

\noindent \textbf{Visual Stimuli ( representing Heritage and Contemporary life).} We use visual artworks from WikiArt~\cite{wikiart2020} to ground conversations between two agents. The diversity of artworks in terms of periods, art genres (landscape, portrait, still life, etc.), and styles (abstract, baroque, cubism, impressionism, etc.) naturally allow for engaging and distinct conversations. We decided to use visual art as the basis of our dataset for three primary reasons. First, artworks are often created with the intention of eliciting emotional responses in viewers, making them a natural choice for studying the relationship 
between visual stimuli and emotions. Second, most existing datasets that feature individual emotion attributions and explanations are centered around art (e.g.~\cite{achlioptas2021artemis,mohamed2022okay,mohamed2022artelingo}). Thirdly,
art heritage embodies human experiences, which we aim to showcase through WikiArt’s coverage of centuries of human evolution and expression including 588 contemporary/real-life images (i.e., recent photography). Fig.~\ref{figure:visual_stimuli} shows a few examples from different art genres, including photos.
\noindent \textbf{Data Inclusion and Exclusion} are crucial steps in building any dataset of high quality. In our data inclusion process, we only consider the dialogs that meet our criteria of quality and relevance. Specifically, we include only those dialogs that contain the full 10 turns, where both Questioner and Answerer provide their emotional explanations about the hidden image. Out of \textbf{107,912} dialogs, 17,435 dialogues that had less than 10 turns were deemed ``incomplete" and were excluded from the dataset. We also exclude those dialogs that contain any inappropriate or irrelevant content that does not follow the given instructions such as avoiding offensive messages, chitchatting, etc. The remaining 90,477 complete dialogs were manually inspected, and \textbf{40,477} were excluded for noncompliance with said guidelines. To facilitate a more productive dialogue that delves deeper into the exploration of the hidden image, we required Questioners to avoid redundant questions that can be easily answered by referring to the given opinions. By carefully selecting and filtering the data, we aim to ensure the quality and usefulness of the \datasetshortname{} dataset.

\subsection{Dataset Analysis}

In this section, we analyze our \datasetshortnamenb dataset and compare it to existing similar datasets. In total, our \datasetshortname~dataset contains \datasetsize~dialogs (10 QA pairs) on $50$K unique artworks, resulting in \qapair QA pairs.

\noindent \textbf{Comparison to similar datasets.} Tab.~\ref{tab:analysis} compares our dataset to existing visual dialog datasets. CLEVR~\cite{kottur2019clevr}, MNIST~\cite{seo2017visual} contain more dialogs. However,  they are generated synthetically for the purpose of diagnosis of the Visual Dialog models and do not capture the real natural conversation. Compared to VisDial~\cite{das2017visual}, our dataset contains emotion labels and corresponding explanations for 50k dialogs. Additionally, our dataset features longer questions and answers with a vocabulary size of 21K, three times larger than VisDial. The mean question and answer lengths in AffectVisDial are 10.92 and 14.85, respectively, compared to 5.1 and 2.9 in VisDial. 
\begin{wraptable}{r}{0.45\textwidth}
\begin{adjustbox}{width=0.45\textwidth,center}
\begin{tabular}{ccccccc}
\toprule   Dataset & Type & Nouns  &  Pronouns   & Adjectives  &  Adpositions   & Verbs \\ 
 \midrule
VisDial~\cite{das2017visual} & Q & 1.4 & 0.6 & 0.3 & 0.3 & 1.4\\
\datasetshortnamenb{} (Ours) & Q & \textbf{2.6} & \textbf{1.0} & \textbf{0.5} & \textbf{1.4} & \textbf{2.1}\\
\addlinespace[1pt]
\cdashline{1-7}[1.5pt/2pt]
\addlinespace[1pt]
VisDial~\cite{das2017visual} & A & 0.9 & 0.2 & 0.3 & 0.2 & 0.5\\
\datasetshortnamenb{} (Ours) & A & \textbf{3.6} & \textbf{1.0} & \textbf{1.1} & \textbf{1.9} & \textbf{2.7}\\
\midrule
\multirow{8}{*}{ \datasetshortnamenb{} (Ours)} 
& Positive Q & 2.6 & 1.1 & 0.5 & 1.4 & 2.0\\
& Negative Q & 2.6 & 0.9 & 0.5 & 1.4 & 2.0\\
& Positive A & 3.5 & 0.9 & 1.1 & 1.8 & 2.6\\ 
& Negative A & 3.4 & 1.0 & 1.1 & 1.8 & 2.6\\
& Qers' Exp  w/o image & 5.9 & 2.6 & 2.3 & 3.3 & 5.8 \\
& Qers' Exp  w/ image & 3.9 & 2.0 & 1.61 & 2.4 & 4.0 \\
& Aers' Exp & 6.3 & 2.8 & 2.6 & 3.4 & 6.1 \\
%& Exp in total & 14.2 & 3.9 & 5.4 & 7.1 & 12.4\\
\bottomrule
\addlinespace[2pt]
\end{tabular}
\end{adjustbox}
\caption{ Language richness reported as the average Part-of-Speech units per individual Questions (Q), Answers (A), and Explanations (Exp) from Questioners (Qers) and Answers (Aers).}
\label{tab:pos_analysis} 
\vspace{-4mm}
\end{wraptable}

\noindent \textbf{Linguistic analysis.} Tab.~\ref{tab:pos_analysis} presents the Part-of-Speech (PoS) analysis for questions and answers in \datasetshortnamenb{} and VisDial. Compared to VisDial, \datasetshortnamenb{} has a higher frequency of PoS tags in questions and answers, suggesting that the answers are more detailed and descriptive of the image attributes and activities. Emotional explanations in \datasetshortnamenb{} have higher PoS tags than questions and answers. This is expected since the agents should be expressive in their explanations in order to convey what constructed their emotion from the dialog. We also break down the results for dialogs concluded with a positive emotion (i.e., amusement, awe, excitement, contentment) $vs$ negative emotion (i.e., sad, anger, disgust, fear).  We found that there are not many statistical differences between questions from dialogs with positive conclusions and questions from dialogues with negative conclusions.

\begin{figure}[!tbp]
\begin{adjustbox}{width=\linewidth,center}
    \centering
    \includegraphics{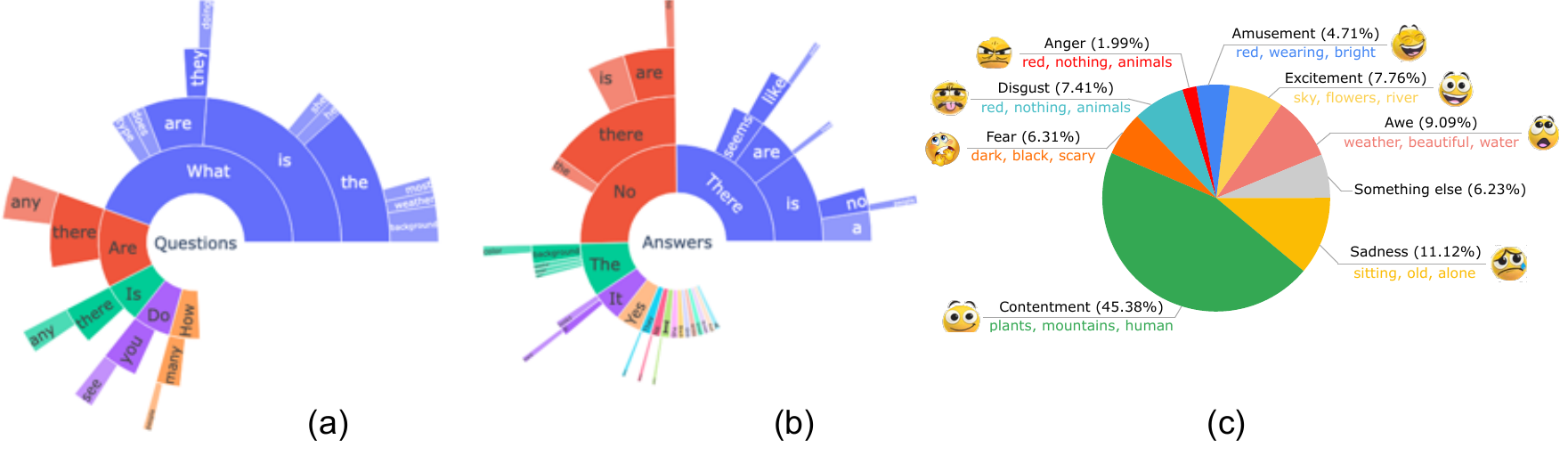}
\end{adjustbox}
    \caption{Distribution of first n-grams for \datasetshortname{} (a) questions, (b) answers. (c)  Questioner's emotion distribution before observing the hidden image, along with the most affective words for each specific emotion from the dialogue.}
    \label{fig:sunburst}
\end{figure}

\begin{table}[!t]
    \centering
\begin{adjustbox}{width=0.75\linewidth,center}
\begin{tabular}{ccccc}
\toprule   \textbf{Datasets}  & CLEVR Dialog~\cite{kottur2019clevr} & MNIST Dialog~\cite{seo2017visual}  &  VisDial~\cite{das2017visual} & \datasetshortname{} \textbf{(Ours)} \\ 
 \midrule
Type (real-R, synthetic-S) & S & S & R & R  \\
\# Images &85K &50K &123K & 50K \\
\# Dialogs &425K &150K &123K & \dialognum  \\
\# Questions &4.25M &1.5M &1.2M & \qnum  \\
\# Unique Q &73K &335 &380K & \uniqueq  \\
\# Unique A &29 &38 &340K & \uniquea  \\
Vocab. Size & 125 & 54 & 7K & \vocab  \\
Mean Q Len &10.6 & 8.9 & 5.1 & \qlen  \\
Mean A Len & 1 & 1 & 2.9 & \alen \\
\midrule
Mean Q Explanation Len & NA & NA & NA & \explenbefore   \\
Mean A Explanation Len & NA & NA & NA & \explenafter  \\
Emotion labels & NA & NA & NA & \dialognum $\times 2$\\
Emotion Explanations & NA & NA & NA & {\dialognum  $\times 2$} \\
\bottomrule
\addlinespace[2pt]
\end{tabular}
\end{adjustbox}
\caption{Comparison with existing visual dialog datasets.}
\label{tab:analysis} 
\vspace{-8mm}
\end{table}

\begin{wrapfigure}{r}{0.35\textwidth}
\begin{center}
      \vspace{-12mm}\includegraphics[width=0.35\textwidth]{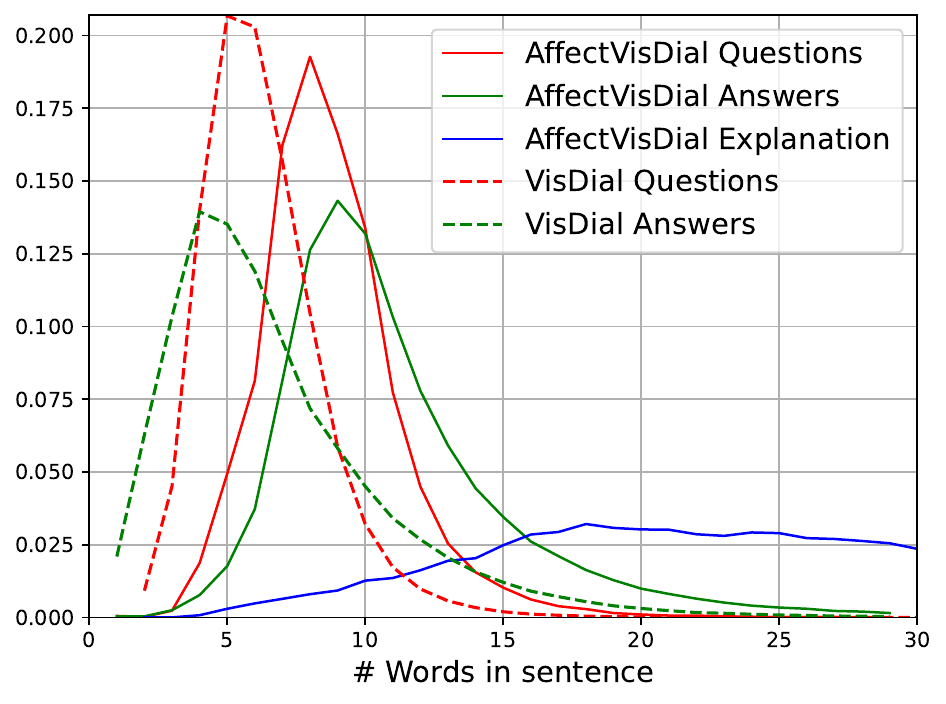}
      \caption{Distribution of lengths for questions, answers, and explanations} \label{figure:distribution}
  \smallskip\par
  \includegraphics[width=0.35\textwidth]{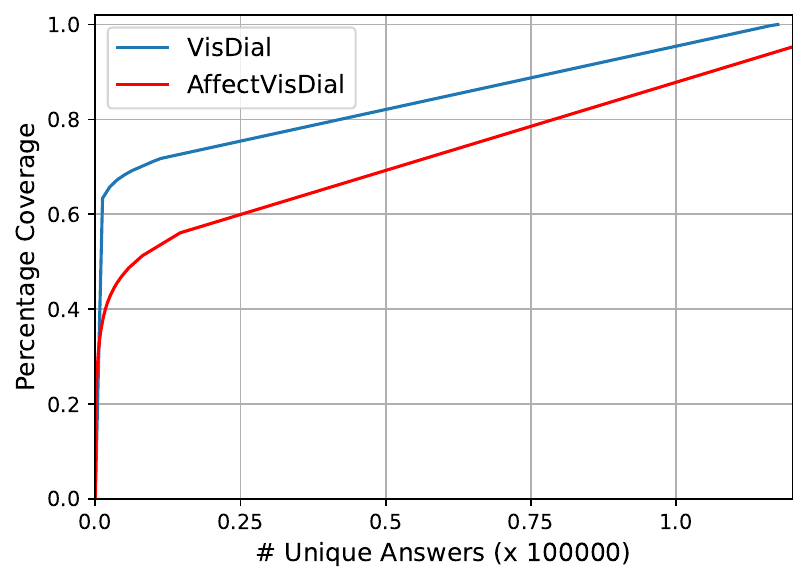}
  \caption{percent coverage of overall answers from the training set covered top frequent unique answers, compared to VisDial. Based on the coverage, \datasetshortnamenb{}'s answers are more diverse.} \label{figure:coverage}
\end{center}
\vspace{-8mm}
\end{wrapfigure}

\noindent \textbf{Analyzing Questions. }
Fig.~\ref{figure:distribution} shows the distribution of the lengths of questions for   \datasetshortnamenb{} and VisDial (see the solid red lines). The plot indicates that questions and answers of \datasetshortnamenb{} are significantly longer. %In Table~\ref{tab:analysis}, on average, questions in \datasetshortnamenb{} are of 8.3 lengths meanwhile, VisDial has 5.1. 
Fig.~\ref{fig:sunburst} (a) shows Sunbursts visualization of the distribution of questions based on the first four words in the dataset.
\emph{Binary Questions vs Binary Answers:} It is interesting to see that the top 10 answers are binary answers with details, e.g., \textit{``No there are no trees or plants in the image''} or \textit{``Yes there is a man in the image.''} This is a consequence of the questioner not being able to see the image. Thus, they are asking contextually relevant questions like \textit{``Do you see any trees or plants in the image?''} in an attempt to explore the scene. Das et al. \cite{das2017visual} observed that the VQA dataset contains binary questions that can be answered by simply 
``\textit{yes}"/``\textit{no}"/``\textit{maybe}". Regarding answers which contain  ``\textit{yes}"/``\textit{no}" as binary answers, they reported that only 46.96\% of all  ``\textit{yes}"/``\textit{no}" responses were binary answers containing  ``\textit{yes}" in VisDial. However, in VQA there was a bias towards  ``\textit{yes}", with 61.40\% of binary responses being  ``\textit{yes}". In our dataset, there is a balance between binary questions containing  ``\textit{yes}" and ``\textit{no}", 50.32\% and 49.68\%, respectively. The percentage of ``\textit{yes}"/``\textit{no}" questions with respect to the total number of answers from the training set is   52.7\% for VisDial and for 
 25.5\% for \datasetshortnamenb{} ($>$ 50\%  reduction).

\noindent \textbf{Analyzing Answers.}
Answers in \datasetshortnamenb are much longer and more informative - average length \alen words; see Tab.~\ref{tab:analysis} and the solid green and dashed green plots in Fig.~\ref{figure:distribution}. The  top-1000 frequent answers in VisDial~\cite{das2017visual} cover 63\% of all answers, while for \datasetshortnamenb they cover only \answertopcover.  Since the number of dialogs in VisDial is higher than \datasetshortnamenb{}, we sampled a subset of dialogs of the same size as our whole dataset size and computed the percentage coverage of answers on this subset from VisDial. Fig.~\ref{figure:coverage} shows the percent coverage of unique answers overall answers in the training set. 
Fig.~\ref{fig:sunburst} (b) visualize answer distribution based on the starting few words. A large portion of answers are descriptive to convey the existence of objects in the image (e.g.,  ``\textit{there are} / \textit{is}" or ``\textit{there seems like}"), and many answers start with``\textit{there are}" or ``\textit{is}" since questioners are querying whether things are in the image.

\noindent \textbf{Analyzing emotion explanations.}
We report three types of analysis for explanations: explanations of the Questioner (1) before and (2) after the image observation, and Answerer explanations. Tab.~\ref{tab:analysis} shows that on average explanations have lengths  27.29 and 28.95 for Questioner and Answerer, respectively. Tab.~\ref{tab:pos_analysis} shows that the average number of PoS from explanations before observing images is larger than those after observing the image. Answerer explanation contains slightly more pronouns, so we speculate that they can easily refer to visual cues since they can see the hidden image.

\begin{wrapfigure}{l}{0.5\textwidth}
    \begin{center}
        \vspace{-10mm}\includegraphics[width=0.5\textwidth]{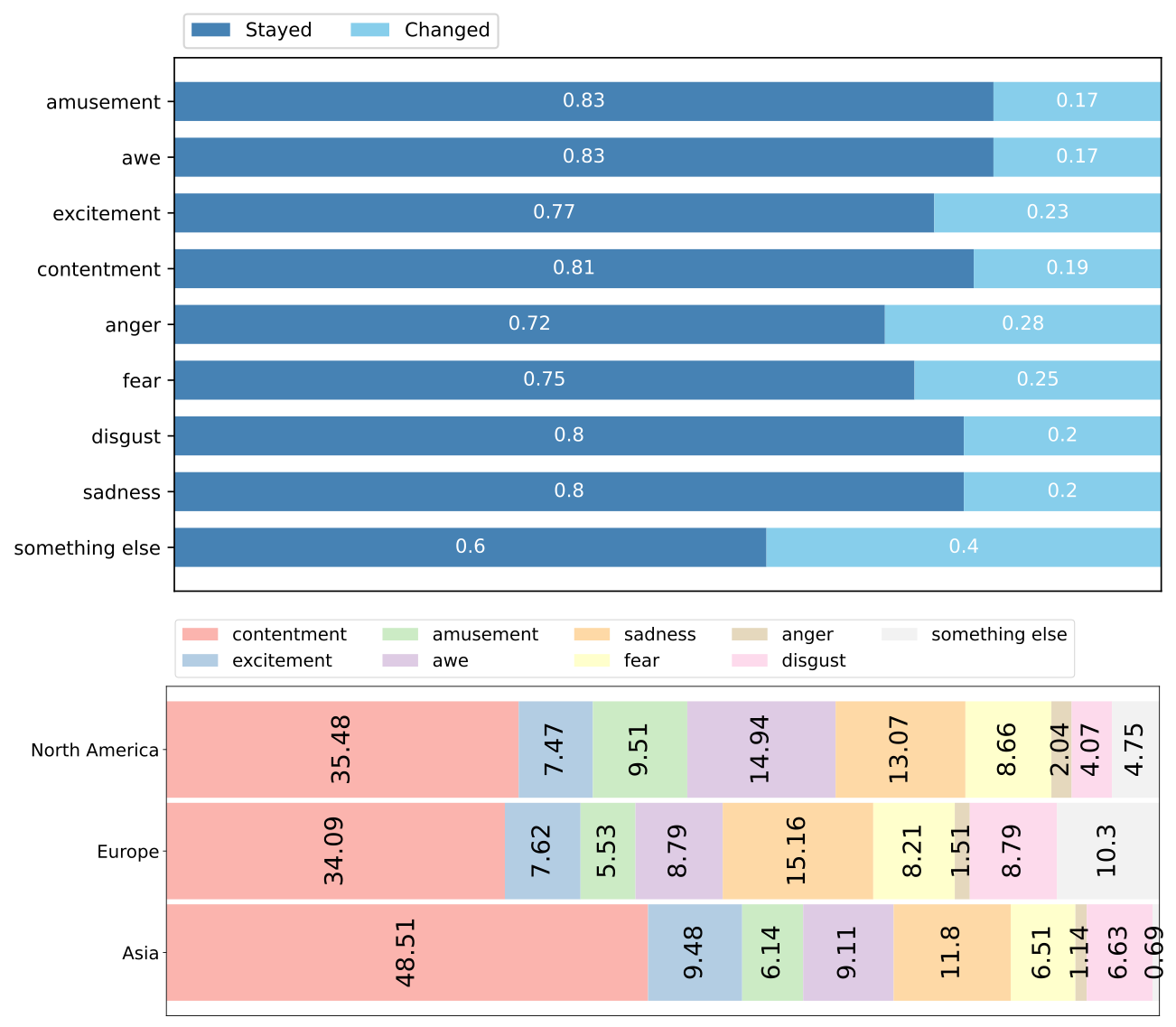}
        \caption{\textbf{Top:} percentage of change of emotion before/after the Questioner observes the image in the dataset. \textbf{Bottom:} emotion distribution across continents.}
        \label{figure:emotion_change}
        \vspace{-6mm}
    \end{center}
\end{wrapfigure}

\noindent \textbf{Emotion Distribution.} 
We report emotion distribution before observing the image in Fig.~\ref{fig:sunburst} (c) with the top words associated with each emotion. It is obvious that the most dominant emotion is \textit{``contentment"}. Fig.~\ref{figure:emotion_change} (top) shows the percentage of Questioner, who changed their emotion choice after observing the hidden image. On average, 23.22\% of the time, annotators changed their minds after looking at the hidden image. In general, dialog information is informative enough for the Questioner to conclude their final emotional responses. The emotion category that was influenced the most by the emotion change is \textit{``something else''}, where the Questioner may not have to find enough linguistic details associated with images.
Fig.~\ref{figure:emotion_change} (bottom) shows the distribution of emotions before observing the image across continents; \textit{``contentment"} is the most dominant emotion across annotators from all continents. Interestingly, we can observe that there are some similarities in the emotion distribution between North America and Europe: the emotion labels like \textit{``contentment"}, \textit{``excitement"}, and\textit{``fear"} are almost identical between these groups.

\section{Task and Neural Baselines}
\label{sec:modeling}

The \taskname{} task involves three subtasks: dialog-based question answering~\ref{sec:dial_qa}, affective explanation generation~\ref{sec:exp_gen} and dialog-based emotion classification~\ref{section:dial_classification}.  We split the dataset into  train,
validation, and test sets with 80\%, 5\%, 15\% percentages, respectively.

\noindent \textbf{Notations:}
To begin with, we first elaborate on several notations. Specifically, $I$ represents an image, and two emotion labels (in the form of positive-negative pairs), denoted as $E$, and associated opinions as $C$. Moreover, a dialog $D$ containing 10 pairs of questions and answers about the image can be represented as $D = \{(Q_1, A_1), ... , (Q_{10}, A_{10})\}$. At any given round $t$, a context $H_t$ is defined as the combination of the image, emotion labels, captions, and the dialog history up to that point, which can be represented as $H_t = \{I, C, (Q_1, A_1), ... , (Q_{t-1}, A_{t-1})\}$.
% \subsection{\taskname{} Skills}

\subsection{Dialog-based Question Answering Subtask}
\label{sec:dial_qa}
In this subtask, neural systems' aim is to generate an answer $A_t$ for a given question $Q_t$ and context $H_t$ at round $t$, similar to \cite{das2017visual}. Two approaches can be used: generative~\cite{nguyen2020efficient, chen2022utc, murahari2019improving} and discriminative~\cite{chen2022utc, murahari2020large, nguyen2020efficient}. Generative models generate an answer, while discriminative models rank pre-defined candidate answers. Candidates are generated following~\cite{das2017visual}, including a ground truth answer, 50 possible answers, 30 most popular answers, and 19 random answers.

\noindent \textbf{Neural Baselines:} We explore several neural models for this task. Firstly, we utilize the simple Nearest Neighbors approach, as introduced in Das et al.~\cite{das2017visual}. We experiment with two variants: NN-Q, which finds $k$ similar questions based on GloVE~\cite{pennington2014glove} embeddings and takes the average similarity scores between candidates and $k$ answers for ranking options; and NN-QI, which extracts $K$ similar questions and then selects a subset of $k$ questions based on image similarity features from VGG-16~\cite{simonyan2014very}. Additionally, we evaluate two state-of-the-art models: VisDial-BERT~\cite{murahari2020large}, which adopts VilBERT~\cite{lu2019vilbert} pretrained on Conceptual Captions~\cite{sharma2018conceptual} and VQA~\cite{antol2015vqa} and is fine-tuned on Visual Dialog~\cite{das2017visual} tasks; and LTMI~\cite{nguyen2020efficient}, a SOTA model that can be used as both a generative and discriminative model.

\subsection{Affective Explanation Generation  Subtask}
\label{sec:exp_gen}

In this subtask, we investigate the ability of neural models to generate emotion explanations given $D$ and, optionally $I$. Additionally, we incorporate $E$ and $C$ to provide a comprehensive context for the generation process. To simultaneously generate the emotion class and its corresponding explanation for both the questioner and answerer, we experiment with different generative language models. To make the explanation more faithful to the identified emotion, we design a prompt ``\textit{I feel \textbf{EMOTION} because \textbf{EXPLANATION}}'' and train the models with this prompt. Generating emotion and explanation together helps the model understand the reasoning behind the emotion and produces more accurate predictions. 

\noindent \textbf{Neural Baselines:}
We consider two generative text models (BART~\cite{lewis2020bart}, T5-large~\cite{raffel2020exploring}) and a recently introduced multi-modal model NLX-GPT~\cite{sammani2022nlx} for emotion and explanation generation for both Questioner and Answerer. Since the Answerer has always access to the image $I$, we include $I$ in the form of text from pretrained BLIP~\cite{li2022blip} model for text models T-5 and BART. For the Questioner, we experiment with two variants: the model (1) w/o and (2) w/ access to the image $I$. For (2), we also include $I$ in the input similar to the Answerer case and predict the emotion and explanation of the Questioner after observing the image. We fine-tune the NLX-GPT model~\cite{sammani2022nlx} for generating emotional explanations, which can accept both visual and language inputs and produce answers and explanations. We train the model from the perspectives of both the Questioner and the Answerer using the same input as the previously mentioned language models. The image $I$ is passed through the model's image encoder before being fed to the decoder for final emotion and explanation generation (see the supplementary for implementation details).
We also report the performance of recent Large Language Models (LLMs) and vision LLMs, specifically LLaMa2-7b~\cite{touvron2023llama} and MiniGPT-4-v2~\cite{chen2023minigptv2} (based on LLaMa2) in a zero-shot and instruction fine-tuning setup.

\subsection{Dialog-based Emotion Prediction Task}
\label{section:dial_classification}
We propose a dialog-based prediction task that aims to predict the emotion label based on a visually grounded conversation, which can be formulated as a standard document classification problem with 9 emotion categories. To address this task, we utilize a pre-trained RoBERTa~\cite{liu2019roberta} model and fine-tune it on our dataset. As an input for this model, we provide two opinions $C$ and dialog $D$.

\section{Experimental Results}

\noindent \textbf{Evaluation Metrics.}
For the Dialog-based Question Answering task, we use standard retrieval-based metrics introduced in \cite{das2017visual}: mean rank (MR), mean reciprocal rank (MRR), and recall R@$k$ ($k = 1, 5, 10$) based on 100 candidate answers. Candidates include a ground truth answer, 50 possible answers, 30 popular answers, and 19 random answers. 
\begin{wraptable}{r}{0.45\textwidth}
\vspace{-10mm}
\begin{adjustbox}{width=0.45\textwidth,center}
\begin{tabular}{llllll}
\toprule
Model & R@1($\uparrow$) & R@5($\uparrow$) & R@10($\uparrow$) & MRR($\downarrow$) & MR($\downarrow$) \\
\midrule
NN-Q  &   0.11  &  0.26  &  0.38    &   0.65  &  32.01  \\
NN-QI  &   0.07  &   0.16  &  0.31    &  0.71  &  36.32  \\
LTMI-G~\cite{nguyen2020efficient} &   0.18 &   0.25  &  0.55    &   0.36  &  24.47 \\
LTMI-D~\cite{nguyen2020efficient} &   0.25  &   0.39  &  0.84   &   0.58  &  5.62 \\
VisDial-BERT~\cite{murahari2020large} &   0.23  &   0.35  &  0.79    &   0.47 &  6.10  \\
\bottomrule
\end{tabular}
\end{adjustbox}
\caption{Results on dialog-based question answering task.}
\label{table:visdial-results}
\vspace{-5mm}
\end{wraptable}To evaluate explanation generation quality and linguistic similarity to human explanations, we use popular machine-based captioning metrics: BLEU~\cite{papineni2002bleu}, BERT-score~\cite{zhang2019bertscore}, and BART-score~\cite{yuan2021bartscore}. We evaluate predicted emotion categories using a weighted F1-score due to data imbalance over the 9 emotion classes.

\noindent \textbf{Dialog-based Emotion Prediction.} After fine-tuning our RoBERTa-based classifier, we observe that it achieves 62.2\% accuracy. It's important to acknowledge the inherent challenge in emotion prediction due to its subjective nature.

\noindent \textbf{Dialog-based Question Answering.}
Tab.~\ref{table:visdial-results} shows the results of the dialog-based question-answering subtask, where LTMI-D achieves the highest scores among all baselines in terms of retrieval metrics. Naive neighboring approaches perform poorly, possibly due to the large number of unique answers compared to questions. Discriminative methods such as VisDial-BERT and LTMI-D outperform the generative LTMI-G on average by 0.06, 0.12, and 0.27 in R@1, R@5, and R@10 metrics, respectively. The results suggest that the task is challenging in its open form, and generative models may not be as effective as discriminative models.

\noindent  \textbf{Questioner Explanation Generation.}
Tab.~\ref{table:explanation-results-questioner} presents the results for the explanation generation subtask. Models relying solely on opinions ($C$) and emotion representations ($E$) as input exhibit significantly poorer performance across all metrics compared to other baselines. For instance, the BART model without dialogue ($D$) underperforms its counterpart with dialogue by 0.24 BLEU scores, indicating the insufficiency of relying solely on two input opinions. Incorporating textual descriptions of images ($I$) for text-based models leads to noteworthy improvements, as they provide valuable information \begin{wraptable}{r}{0.45\textwidth}
    \vspace{-4mm}
    \caption{Results on Affective Explanation Generation  setup for Questioner. $I,E,C,D$ represents the image, 2 opposed emotion labels, associated opinions, and the dialog defined in Sec.~\ref{sec:modeling}.}
\label{table:explanation-results-questioner}
\begin{adjustbox}{width=0.45\textwidth,center}
\begin{tabular}{c|c|c|c|c|c|c|c|c}
\toprule
     Model & $I$ & $E$ & $C$ & $D$ & BLEU($\uparrow$) & BERT($\uparrow$) & BART($\downarrow$) & Emo-F1($\uparrow$) \\
     \hline
     NLX-GPT~\cite{sammani2022nlx} & $\times$ & \checkmark & \checkmark & $\times$ & 0.09 & 0.65 & -6.67 & 29.55 \\
    NLX-GPT~\cite{sammani2022nlx} & $\times$ & \checkmark & \checkmark & \checkmark & 0.23 & 0.86 & -5.05 & \textbf{46.17} \\
    \addlinespace[1pt]
    \cdashline{1-9}[1.5pt/2pt]
    \addlinespace[1pt]
    NLX-GPT~\cite{sammani2022nlx} & \checkmark & \checkmark & \checkmark & $\times$ & 0.11 & 0.70 & -5.72 & 35.50 \\
    NLX-GPT~\cite{sammani2022nlx} & \checkmark & \checkmark & \checkmark & \checkmark & \textbf{0.27} & \textbf{0.88} & \textbf{-4.92} & 43.61 \\
    \midrule
    BART-Large~\cite{lewis2020bart} &  $\times$ & \checkmark & \checkmark & $\times$ &0.02 & 0.02 & -5.71 & 2.10 \\
    BART-Large~\cite{lewis2020bart} &  $\times$ & $\times$& \checkmark & \checkmark & 0.26 & 0.88 & -4.25 & 47.49\\
    BART-Large~\cite{lewis2020bart} &  $\times$ & \checkmark & \checkmark & \checkmark & 0.27 & 0.88 & -4.26 & \textbf{51.62}\\
    \addlinespace[1pt]
    \cdashline{1-9}[1.5pt/2pt]
    \addlinespace[1pt]
    BART-Large~\cite{lewis2020bart} &  \checkmark & \checkmark & \checkmark & $\times$ & 0.04 & 0.85 & -5.38 & 38.11 \\
    BART-Large~\cite{lewis2020bart} &  \checkmark & $\times$ & \checkmark & \checkmark & 0.27 & 0.88 & -4.18 & 45.52 \\
    BART-Large~\cite{lewis2020bart} &  \checkmark & \checkmark & \checkmark & \checkmark & \textbf{0.28} & \textbf{0.89} & \textbf{-4.18} & 50.43\\
    \midrule
    T5-Large~\cite{raffel2020exploring} &  $\times$ & \checkmark & \checkmark & $\times$ & 0.05 & 0.05 & -5.98 & 15.20 \\
    T5-Large~\cite{raffel2020exploring}  &  $\times$ & $\times$ & \checkmark & \checkmark & 0.25 & 0.88 & -4.54 & 40.48\\
    T5-Large~\cite{raffel2020exploring}  &  $\times$ & \checkmark & \checkmark & \checkmark & 0.26 & 0.87 & -4.4 & \textbf{46.34} \\
    \addlinespace[1pt]
    \cdashline{1-9}[1.5pt/2pt]
    \addlinespace[1pt]
    T5-Large~\cite{raffel2020exploring} &  \checkmark & \checkmark & \checkmark & $\times$ & 0.04 & 0.85 & -5.38 & 38.11 \\
    T5-Large~\cite{raffel2020exploring} &  \checkmark & $\times$ & \checkmark & \checkmark & 0.28 & 0.88 & -4.35 & 41.91 \\
    T5-Large~\cite{raffel2020exploring} &  \checkmark & \checkmark & \checkmark & \checkmark & \textbf{0.28} & \textbf{0.89} & \textbf{-4.34} & 42.95\\
    \bottomrule
\end{tabular}
\end{adjustbox}
\end{wraptable} about the visual input. Notably, the T5 model demonstrates considerable enhancements in metrics such as BLEU, BERT, and BART when exposed to image textual descriptions during training and testing. Intriguingly, when dialogues ($D$) are introduced as an additional input for these models, their performance significantly improves across all metrics. For instance, T5-Large and NLX-GPT which incorporate $D$ outperform those that do not take $D$ as input by 0.24 and 0.18 BLEU scores, respectively, further underscoring the critical role of dialogue in guiding the model's emotion-related tasks. The results for Answerer are available in the supplementary.

\begin{wrapfigure}{l}{0.55\textwidth}
    \begin{center}
        \includegraphics[width=0.55\textwidth]{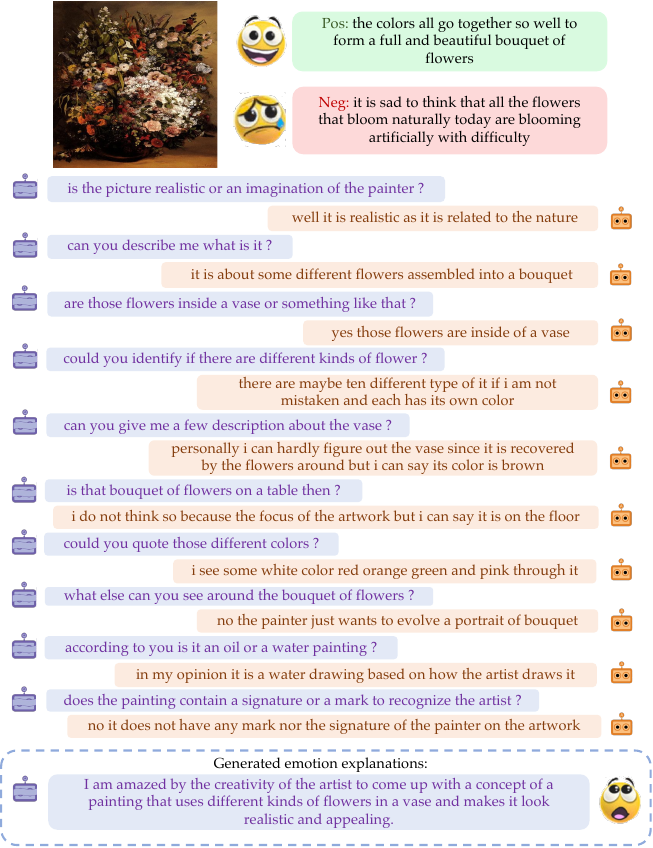}
        \caption{ Generated Questioner's Explanation from our T5-Large baseline.}
    \label{fig:qualitative_output}
    \vspace{-10mm}
    \end{center}
\end{wrapfigure}

\noindent \textbf{Human Studies}. To validate our models' effectiveness, we selected our top-performing model (BART) and conducted studies on Amazon MTurk. We asked participants to evaluate the reasonableness of our dialog-based QA task and conducted a Turing test for emotion explanation generation. Results indicate that over 90\% of produced answers were considered \textit{reasonable} and 55\% of explanations were considered \textit{human-like}, showing the effectiveness of our model in capturing emotional reasoning in visually grounded conversations. Fig.~\ref{fig:qualitative_output} shows the generated emotion and corresponding explanation from the Questioner. The output closely resembles human explanations (more examples are in the supplementary).

\begin{wraptable}{r}{0.45\textwidth} 
\vspace{-5mm}
    \begin{adjustbox}{width=0.45\textwidth,center}
    \begin{tabular}{c|c|c|c|c|c|c|c|c}
    \toprule
     Model & $I$ & $E$ & $C$ & $D$ & BLEU($\uparrow$) & BERT($\uparrow$) & BART($\downarrow$) & Emo-F1($\uparrow$) \\
     \hline
        LLaMA2-7b (zs)&  $\times$ & \checkmark & \checkmark & $\times$ & 0.02 & 0.62 & -5.29 & 2.49\\
        LLaMA2-7b (zs)&   $\times$ & $\times$ & \checkmark & \checkmark & 0.018 & 0.81 & -5.18 & 5.25\\
        LLaMA2-7b (zs)&  $\times$ & \checkmark & \checkmark & \checkmark & \textbf{0.02} & \textbf{0.85} & \textbf{-5.12} & \textbf{14.45}\\
            \cdashline{1-9}[1.5pt/2pt]
        GPT-4 (zs)  & $\times$ & \checkmark & \checkmark & $\times$  & 0.018 & 0.81 & -5.12 & 24.28 \\
        GPT-4 (zs) & $\times$ & \checkmark & \checkmark & \checkmark  & \textbf{0.021} & \textbf{0.83} & \textbf{-5.11} & \textbf{29.79} \\
        \midrule
        MiniGPT-4-v2 (zs) &  \checkmark & \checkmark & \checkmark & $\times$ & 0.009 & 0.84 & -5.15 & 23.46\\
        MiniGPT-4-v2 (zs) &   \checkmark & \checkmark & \checkmark & \checkmark & \textbf{ 0.01} & \textbf{0.85} & \textbf{-5.14} & \textbf{25.28}\\

        \hline
        
        LLaMA2-7b (ft) &  $\times$ & \checkmark & \checkmark & $\times$ & 0.01 & 0.84 & -5.12 & 28.92\\
        LLaMA2-7b (ft) &   $\times$ & $\times$ & \checkmark & \checkmark & 0.043 & 0.86 & -4.92 & 32.64\\
        LLaMA2-7b (ft) &  $\times$ & \checkmark & \checkmark & \checkmark & \textbf{0.052} & \textbf{0.86} & \textbf{-4.58} & \textbf{37.02} \\

        \midrule
        MiniGPT-4-v2 (ft) &  \checkmark & \checkmark & \checkmark & $\times$ & 0.005 & 0.83 & -5.14 & 36.53\\
        MiniGPT-4-v2 (ft) &   \checkmark & \checkmark & \checkmark & \checkmark & \textbf{0.04} & \textbf{0.86} & \textbf{-4.61 }& \textbf{46.62}\\
        \bottomrule
        
    \end{tabular}
    \end{adjustbox}
    \caption{Results on  Affective Explanation Generation in zero-shot setting and fine-tuned setting. (zs) denotes zero-shot, and (ft) denotes fine-tuned.}
\label{table:explanation-results-zero-shot}
\vspace{-5mm}

\end{wraptable} \noindent \textbf{Zero-shot and Fine-tuned Performance using LLMs and Vision-LLMs.} We explored the potential of multimodal and language foundation models, known for their impressive zero-shot question-answering performance, for predicting emotions and generating corresponding explanations on our newly proposed dataset. Specifically, we evaluated the performance of LLaMa2-7b~\cite{touvron2023llama}, the latest API of GPT-4~\cite{chatgptapi}, and recently released MiniGPT-4-v2~\cite{chen2023minigptv2} with our designed prompts, as inputs to enable these models to generate emotions and their explanations. Table~\ref{table:explanation-results-zero-shot} presents results for models in zero-shot emotion explanation generation. The trend observed indicates that incorporating dialogs as input improves the results. Emotion F1 scores improved from 23.46 to 25.28 and from 24.28 and 29.79 for MiniGPT-4-v2 and GPT-4, respectively. Despite being powerful models trained on massive data, their performance lags behind our trained baselines, suggesting the need for considering emotional alignment with humans. For example, our baselines trained on our dataset showed superior results, yielding more than $40$ in F1 score. These limitations highlight the significance of our dataset in advancing emotion-aware AI systems for future applications. In addition, we fine-tuned the open-source LlaMa2-7b~\cite{touvron2023llama} and MiniGPT-4-v2~\cite{chen2023minigptv2} models on our dataset using instruction fine-tuning and assessed their performance. Table~\ref{table:explanation-results-zero-shot} shows that fine-tuning enhances the performance of these models, as evidenced by higher Emotion F1 scores. This outcome underscores the significance of our dataset in improving model understanding and generation capabilities in terms of affective explanation. Further details about prompts and generated examples are provided in the supplementary.
\section{Discussions}

In this section, we discuss and demonstrate potential AI skills that may benefit from our benchmarks and models. We also summarize our key conclusions:

\begin{figure}[t!]
\begin{adjustbox}{width=1.0\linewidth,center}
    \centering
    \includegraphics{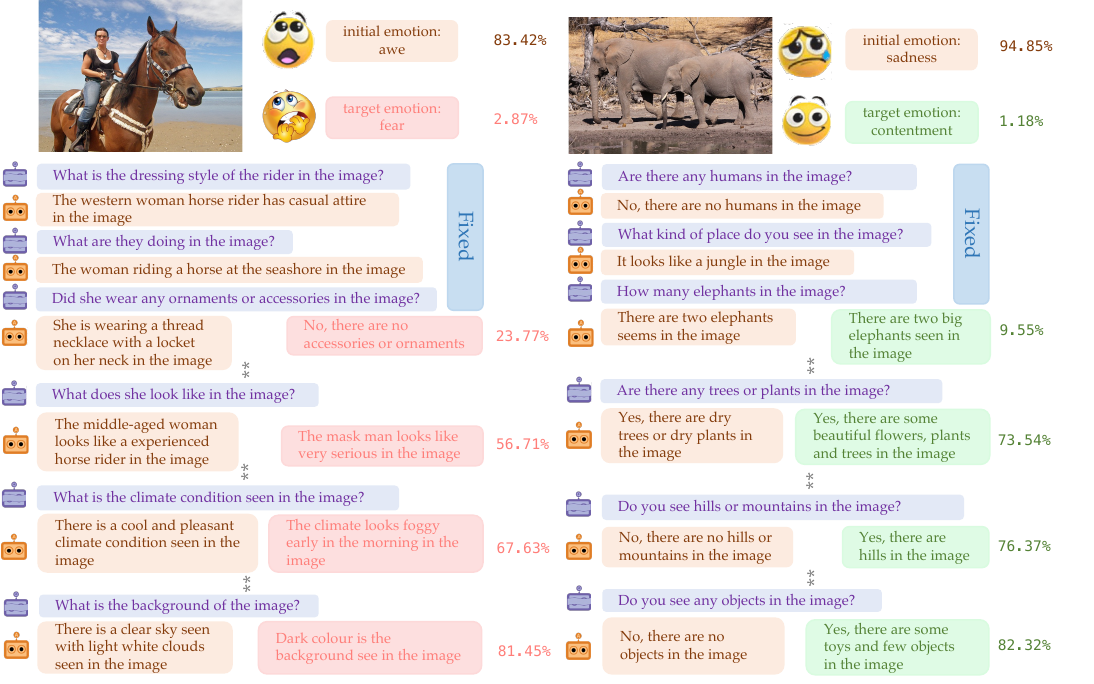}
\end{adjustbox}
    \caption{Examples of altering answers to evoke opposite emotions. The left side of each example shows the original dialog and the right side shows the replaced answer with the corresponding probability of the target emotion}
    \label{fig:emo_alter}
%\vspace{-2mm}
\end{figure}

\noindent \textbf{Emotion Guided Answer Generation.} 
An intriguing avenue for exploration is whether answers can be tailored to elicit specific emotions. In this study, we examine the impact of dialogue answers on resulting emotions by manipulating answers to provoke targeted emotional reactions. To achieve this, we select answer candidates for each question that maximize the probability of the target emotion, as determined by our pretrained RoBERTa-based emotion classifier~\cite{liu2019roberta}. The initial two or three turns of the dialogue remain unchanged, while subsequent answers are replaced by selected candidates. Fig.~\ref{fig:emo_alter} illustrates how gradually altering answers can shift the original emotion towards an opposing direction, highlighting the role of linguistic details in influencing emotions (e.g., the absence of trees increases the probability score of \textit{"fear"}).

\begin{figure}
\centering
    \includegraphics[width=0.65\textwidth]{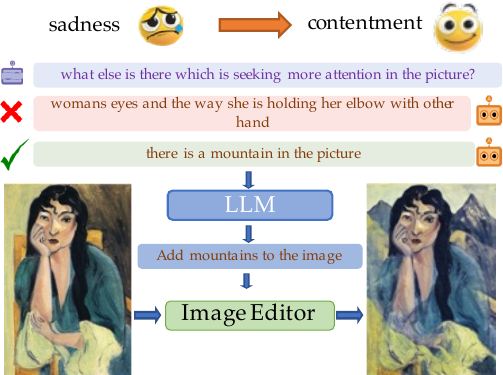}
         \caption{Example of altering answers to evoke opposite emotions, and make \textit{``edit"} in the original image.}
    \label{fig:emo_img_editing}
    \vspace{-8mm}
\end{figure}

\noindent \textbf{Towards Emotionally Reasoned Image  Editing.} While the above study offers valuable insights, it is essential to acknowledge certain limitations. Emotions are complex and subjective, and the classifier's output may not always align perfectly with true human emotional experiences. Additionally, chosen answers may not necessarily be grounded in the visual input. 
An interesting exploration is editing the initial visual input to align with candidate answers based on the selected emotion. We use our RoBERTa-based classifier to choose an answer for a given question, increasing the probability score of the target emotion. The chosen answers are then converted into instructions as input to an image editing model~\cite{Brooks_2023_CVPR} to modify the original image accordingly. Fig.~\ref{fig:emo_img_editing} demonstrates how the emotion \textit{"contentment"} of the visual stimuli can be maximized by suggesting the inclusion of a mountain in the image. This approach paves the way for collaborative image editing systems capable of offering emotionally informed content creation (similar to ~\cite{Weng_2023_ICCV}). While our approach is a preliminary step, future methods may consider using Reinforcement Learning to build a reward function for increased alignment of emotions and text-to-image content~\cite{black2023ddpo, li2018a2, bai2022training, lee2023aligning}.

\noindent \textbf{Automatic Evaluation Challenges.} The subjective nature of this problem, compounded by the use of only a single dialog response per image in our dataset, may lead to limited performance measurements of models. Furthermore, the process of gathering additional references is both costly and complex. Consequently, conventional metrics such as BLEU or BERT, which are reference-based, might inappropriately penalize models. These metrics have demonstrated a relatively low correlation with human judgments, particularly in tasks demanding creativity and diversity. 

Inspired by the G-eval framework~\cite{liu-etal-2023-g}, which was shown to be highly correlated with human judgments, we adopt a reference-free evaluation approach aimed at assessing model performance in terms of coherence, the groundedness of explanations in dialogues, fluency, and the relevance of explanations to emotions. To evaluate this, we sampled 1000 generations from different models for the same input and evaluated using G-eval. Table~\ref{table:g_eval} shows that our fine-tuned VLL/LLMs on our dataset are better than zero-shot counterparts in terms of the groundedness of explanations in dialogues and relevance of explanations to emotions. For example, MiniGPT4-v2 fine-tuned on our dataset achieves 2.04 in groundedness compared to the base version with a 1.89 score. %\textcolor{red}{It is best to prepare also some human evaluation for this part by supplementary. Also make it clear that Groundedness here means grounded to the language in the dialog, not to the image. Do we have any measure for visual grounding understanding?vs }

\begin{table}[!t]
\centering

\begin{adjustbox}{width=0.85\textwidth,center}
\begin{tabular}{cccccc}
\toprule
Method & Coherence & Groundedness & Fluency & Relevance & Overall\\
\hline
LLaMa2-7b (zero-shot) & 2.68 & 2.68 & 3.10 & 2.02 & 2.62 \\
\hline
LLaMa2-7b (fine-tuned) & 2.66 & 2.82 & 2.83 & 2.34 & 2.66 \\
\hline
MiniGPT4-v2 (zero-shot) & 1.77 & 1.87 &  1.89 & 1.37 & 1.73\\
\hline
MiniGPT4-v2 (fine-tuned) & 1.75 & 2.04 & 1.88 & 1.88 & 1.89\\
\hline
NLX-GPT &  2.16 & 1.15 & 2.21 & 1.75 & 1.81\\
%\hline
%Human & 2.68 & 2.71 & 2.91 &  2.19 & 2.62 \\
\bottomrule
\end{tabular}
\end{adjustbox}
\caption{Results on G-Eval metrics.}
\label{table:g_eval}
\vspace{-5mm}
\end{table}

\section{Conclusion}

Emotions are an integral part of human nature.
A deeper comprehension of how emotions emerge from natural conversation can aid in the development of more human-compatible machines. In pursuit of this goal, we introduce a novel dataset, \datasetshortnamenb{}, capturing the dialogue between two agents discussing an image and the resultant emotion they construct from the visually grounded dialogues. By adapting and training the state-of-the-art models on our dataset, we show the potential of AI systems for generating affective explanations grounded in dialogues that are similar to human emotions. We also demonstrated the models trained on \datasetshortnamenb{} can facilitate both emotion-guided answer generation and pave the way toward building and advancing emotionally reasoned collaborative image editing systems.  We hope our dataset and models will help ease future research in affective vision and language.

%\hfill \break

\noindent \textbf{Acknowledgement.} This project is funded by KAUST BAS/1/1685-01-01, SDAIA-KAUST Center of Excellence in Data Science and Artificial Intelligence. The authors express their appreciation to Jack Urbanek, Sirojiddin Karimov, and Umid Nejmatullayev for their valuable assistance in data collection setup. Lastly, the authors extend their gratitude to the diligent efforts of the Amazon Mechanical Turkers, DeepenAI, and SmartOne teams, as their contributions were indispensable for the successful completion of this work.

%\clearpage\mbox{}Page \thepage\ of the manuscript.
%\clearpage\mbox{}Page \thepage\ of the manuscript.
%\clearpage\mbox{}Page \thepage\ of the manuscript.
%\clearpage\mbox{}Page \thepage\ of the manuscript.
%\clearpage\mbox{}Page \thepage\ of the manuscript. This is the last page.
%\par\vfill\par
%Now we have reached the maximum length of an ECCV \ECCVyear{} submission (excluding references).
%References should start immediately after the main text, but can continue past p.\ 14 if needed.
%\clearpage  % TODO REVIEW/FINAL: This \clearpage needs to be removed from both review and camera-ready versions.

\bibliographystyle{splncs04}
\bibliography{main}

\begin{thebibliography}{10}
\providecommand{\url}[1]{\texttt{#1}}
\providecommand{\urlprefix}{URL }
\providecommand{\doi}[1]{https://doi.org/#1}

\bibitem{chatgptapi}
Openai api: Gpt-4. \url{https://platform.openai.com/docs/models/gpt-4} (2023), accessed: November 15, 2023

\bibitem{Achlioptas_2023_CVPR}
Achlioptas, P., Ovsjanikov, M., Guibas, L., Tulyakov, S.: Affection: Learning affective explanations for real-world visual data. In: Proceedings of the IEEE/CVF Conference on Computer Vision and Pattern Recognition (CVPR). pp. 6641--6651 (June 2023)

\bibitem{achlioptas2021artemis}
Achlioptas, P., Ovsjanikov, M., Haydarov, K., Elhoseiny, M., Guibas, L.J.: Artemis: Affective language for visual art. In: Proceedings of the IEEE/CVF Conference on Computer Vision and Pattern Recognition. pp. 11569--11579 (2021)

\bibitem{agrawal2018don}
Agrawal, A., Batra, D., Parikh, D., Kembhavi, A.: Don't just assume; look and answer: Overcoming priors for visual question answering. In: Proceedings of the IEEE conference on computer vision and pattern recognition. pp. 4971--4980 (2018)

\bibitem{antol2015vqa}
Antol, S., Agrawal, A., Lu, J., Mitchell, M., Batra, D., Zitnick, C.L., Parikh, D.: Vqa: Visual question answering. In: Proceedings of the IEEE international conference on computer vision. pp. 2425--2433 (2015)

\bibitem{bai2022training}
Bai, Y., Jones, A., Ndousse, K., Askell, A., Chen, A., DasSarma, N., Drain, D., Fort, S., Ganguli, D., Henighan, T., et~al.: Training a helpful and harmless assistant with reinforcement learning from human feedback. arXiv preprint arXiv:2204.05862  (2022)

\bibitem{black2023ddpo}
Black, K., Janner, M., Du, Y., Kostrikov, I., Levine, S.: Training diffusion models with reinforcement learning (2023)

\bibitem{bogdanov2019mediaeval}
Bogdanov, D., Porter, A., Tovstogan, P., Won, M.: Mediaeval 2019: Emotion and theme recognition in music using jamendo. In: Larson M, Hicks S, Constantin MG, Bischke B, Porter A, Zhao P, Lux M, Cabrera Quiros L, Calandre J, Jones G, editors. MediaEval’19, Multimedia Benchmark Workshop; 2019 Oct 27-30, Sophia Antipolis, France. Aachen: CEUR; 2019. CEUR Workshop Proceedings (2019)

\bibitem{Brooks_2023_CVPR}
Brooks, T., Holynski, A., Efros, A.A.: Instructpix2pix: Learning to follow image editing instructions. In: Proceedings of the IEEE/CVF Conference on Computer Vision and Pattern Recognition (CVPR). pp. 18392--18402 (June 2023)

\bibitem{buechel2022emobank}
Buechel, S., Hahn, U.: Emobank: Studying the impact of annotation perspective and representation format on dimensional emotion analysis. arXiv preprint arXiv:2205.01996  (2022)

\bibitem{chen2022utc}
Chen, C., Tan, Z., Cheng, Q., Jiang, X., Liu, Q., Zhu, Y., Gu, X.: Utc: A unified transformer with inter-task contrastive learning for visual dialog. In: Proceedings of the IEEE/CVF Conference on Computer Vision and Pattern Recognition. pp. 18103--18112 (2022)

\bibitem{chen2023video}
Chen, J., Zhu, D., Haydarov, K., Li, X., Elhoseiny, M.: Video chatcaptioner: Towards the enriched spatiotemporal descriptions. arXiv preprint arXiv:2304.04227  (2023)

\bibitem{chen2023minigptv2}
Chen, J., Zhu, D., Shen, X., Li, X., Liu, Z., Zhang, P., Krishnamoorthi, R., Chandra, V., Xiong, Y., Elhoseiny, M.: Minigpt-v2: large language model as a unified interface for vision-language multi-task learning (2023), \url{https://arxiv.org/abs/2310.09478}

\bibitem{chen2018emotionlines}
Chen, S.Y., Hsu, C.C., Kuo, C.C., Ku, L.W., et~al.: Emotionlines: An emotion corpus of multi-party conversations. arXiv preprint arXiv:1802.08379  (2018)

\bibitem{wikiart2020}
Community: Wiki art. \url{https://www.wikiart.org/} (2020), accessed: 2020-11-06

\bibitem{cowen2020music}
Cowen, A.S., Fang, X., Sauter, D., Keltner, D.: What music makes us feel: At least 13 dimensions organize subjective experiences associated with music across different cultures. Proceedings of the National Academy of Sciences  \textbf{117}(4),  1924--1934 (2020)

\bibitem{das2017visual}
Das, A., Kottur, S., Gupta, K., Singh, A., Yadav, D., Moura, J.M., Parikh, D., Batra, D.: Visual dialog. In: Proceedings of the IEEE conference on computer vision and pattern recognition. pp. 326--335 (2017)

\bibitem{demszky2020goemotions}
Demszky, D., Movshovitz-Attias, D., Ko, J., Cowen, A., Nemade, G., Ravi, S.: Goemotions: A dataset of fine-grained emotions. arXiv preprint arXiv:2005.00547  (2020)

\bibitem{diener2009evolving}
Diener, E., Scollon, C.N., Lucas, R.E.: The evolving concept of subjective well-being: the multifaceted nature of happiness.  (2009)

\bibitem{ekman1992argument}
Ekman, P.: An argument for basic emotions. Cognition \& emotion  \textbf{6}(3-4),  169--200 (1992)

\bibitem{fan2017emo}
Fan, J., Thorogood, M., Pasquier, P.: Emo-soundscapes: A dataset for soundscape emotion recognition. In: 2017 Seventh international conference on affective computing and intelligent interaction (ACII). pp. 196--201. IEEE (2017)

\bibitem{goyal2017something}
Goyal, R., Ebrahimi~Kahou, S., Michalski, V., Materzynska, J., Westphal, S., Kim, H., Haenel, V., Fruend, I., Yianilos, P., Mueller-Freitag, M., et~al.: The" something something" video database for learning and evaluating visual common sense. In: Proceedings of the IEEE international conference on computer vision. pp. 5842--5850 (2017)

\bibitem{goyal2017making}
Goyal, Y., Khot, T., Summers-Stay, D., Batra, D., Parikh, D.: Making the v in vqa matter: Elevating the role of image understanding in visual question answering. In: Proceedings of the IEEE conference on computer vision and pattern recognition. pp. 6904--6913 (2017)

\bibitem{gunes2006bimodal}
Gunes, H., Piccardi, M.: A bimodal face and body gesture database for automatic analysis of human nonverbal affective behavior. In: 18th International conference on pattern recognition (ICPR'06). vol.~1, pp. 1148--1153. IEEE (2006)

\bibitem{hung2021emopia}
Hung, H.T., Ching, J., Doh, S., Kim, N., Nam, J., Yang, Y.H.: Emopia: a multi-modal pop piano dataset for emotion recognition and emotion-based music generation. arXiv preprint arXiv:2108.01374  (2021)

\bibitem{kottur2019clevr}
Kottur, S., Moura, J.M., Parikh, D., Batra, D., Rohrbach, M.: Clevr-dialog: A diagnostic dataset for multi-round reasoning in visual dialog. arXiv preprint arXiv:1903.03166  (2019)

\bibitem{lee2023aligning}
Lee, K., Liu, H., Ryu, M., Watkins, O., Du, Y., Boutilier, C., Abbeel, P., Ghavamzadeh, M., Gu, S.S.: Aligning text-to-image models using human feedback. arXiv preprint arXiv:2302.12192  (2023)

\bibitem{lewis2020bart}
Lewis, M., Liu, Y., Goyal, N., Ghazvininejad, M., Mohamed, A., Levy, O., Stoyanov, V., Zettlemoyer, L.: Bart: Denoising sequence-to-sequence pre-training for natural language generation, translation, and comprehension. In: Proceedings of the 58th Annual Meeting of the Association for Computational Linguistics. pp. 7871--7880 (2020)

\bibitem{li2018a2}
Li, D., Wu, H., Zhang, J., Huang, K.: A2-rl: Aesthetics aware reinforcement learning for image cropping. In: Proceedings of the IEEE conference on computer vision and pattern recognition. pp. 8193--8201 (2018)

\bibitem{li2024diplomat}
Li, H., Zhu, S.C., Zheng, Z.: Diplomat: A dialogue dataset for situated pragmatic reasoning. Advances in Neural Information Processing Systems  \textbf{36} (2024)

\bibitem{li2022blip}
Li, J., Li, D., Xiong, C., Hoi, S.: Blip: Bootstrapping language-image pre-training for unified vision-language understanding and generation. In: ICML (2022)

\bibitem{li2017dailydialog}
Li, Y., Su, H., Shen, X., Li, W., Cao, Z., Niu, S.: Dailydialog: A manually labelled multi-turn dialogue dataset. arXiv preprint arXiv:1710.03957  (2017)

\bibitem{lin2014microsoft}
Lin, T.Y., Maire, M., Belongie, S., Hays, J., Perona, P., Ramanan, D., Doll{\'a}r, P., Zitnick, C.L.: Microsoft coco: Common objects in context. In: European conference on computer vision. pp. 740--755. Springer (2014)

\bibitem{liu2021imigue}
Liu, X., Shi, H., Chen, H., Yu, Z., Li, X., Zhao, G.: imigue: An identity-free video dataset for micro-gesture understanding and emotion analysis. In: Proceedings of the IEEE/CVF Conference on Computer Vision and Pattern Recognition. pp. 10631--10642 (2021)

\bibitem{liu-etal-2023-g}
Liu, Y., Iter, D., Xu, Y., Wang, S., Xu, R., Zhu, C.: {G}-eval: {NLG} evaluation using gpt-4 with better human alignment. In: Bouamor, H., Pino, J., Bali, K. (eds.) Proceedings of the 2023 Conference on Empirical Methods in Natural Language Processing. pp. 2511--2522. Association for Computational Linguistics, Singapore (Dec 2023). \doi{10.18653/v1/2023.emnlp-main.153}, \url{https://aclanthology.org/2023.emnlp-main.153}

\bibitem{liu2019roberta}
Liu, Y., Ott, M., Goyal, N., Du, J., Joshi, M., Chen, D., Levy, O., Lewis, M., Zettlemoyer, L., Stoyanov, V.: Roberta: A robustly optimized bert pretraining approach. arXiv preprint arXiv:1907.11692  (2019)

\bibitem{lu2019vilbert}
Lu, J., Batra, D., Parikh, D., Lee, S.: Vilbert: Pretraining task-agnostic visiolinguistic representations for vision-and-language tasks. Advances in neural information processing systems  \textbf{32} (2019)

\bibitem{machajdik2010affective}
Machajdik, J., Hanbury, A.: Affective image classification using features inspired by psychology and art theory. In: Proceedings of the 18th ACM international conference on Multimedia. pp. 83--92 (2010)

\bibitem{mohamed2022artelingo}
Mohamed, Y., Abdelfattah, M., Alhuwaider, S., Li, F., Zhang, X., Church, K.W., Elhoseiny, M.: Artelingo: A million emotion annotations of wikiart with emphasis on diversity over language and culture. In: Proceedings of the 2022 Conference on Empirical Methods in Natural Language Processing (EMNLP) (2022)

\bibitem{mohamed2022okay}
Mohamed, Y., Khan, F.F., Haydarov, K., Elhoseiny, M.: It is okay to not be okay: Overcoming emotional bias in affective image captioning by contrastive data collection. In: Proceedings of the IEEE/CVF Conference on Computer Vision and Pattern Recognition. pp. 21263--21272 (2022)

\bibitem{mollahosseini2017affectnet}
Mollahosseini, A., Hasani, B., Mahoor, M.H.: Affectnet: A database for facial expression, valence, and arousal computing in the wild. IEEE Transactions on Affective Computing  \textbf{10}(1),  18--31 (2017)

\bibitem{murahari2020large}
Murahari, V., Batra, D., Parikh, D., Das, A.: Large-scale pretraining for visual dialog: A simple state-of-the-art baseline. In: European Conference on Computer Vision. pp. 336--352. Springer (2020)

\bibitem{murahari-etal-2019-improving}
Murahari, V., Chattopadhyay, P., Batra, D., Parikh, D., Das, A.: Improving generative visual dialog by answering diverse questions. In: Inui, K., Jiang, J., Ng, V., Wan, X. (eds.) Proceedings of the 2019 Conference on Empirical Methods in Natural Language Processing and the 9th International Joint Conference on Natural Language Processing (EMNLP-IJCNLP). pp. 1449--1454. Association for Computational Linguistics, Hong Kong, China (Nov 2019). \doi{10.18653/v1/D19-1152}, \url{https://aclanthology.org/D19-1152}

\bibitem{murahari2019improving}
Murahari, V., Chattopadhyay, P., Batra, D., Parikh, D., Das, A.: Improving generative visual dialog by answering diverse questions. arXiv preprint arXiv:1909.10470  (2019)

\bibitem{nguyen2020efficient}
Nguyen, V.Q., Suganuma, M., Okatani, T.: Efficient attention mechanism for visual dialog that can handle all the interactions between multiple inputs. In: European Conference on Computer Vision. pp. 223--240. Springer (2020)

\bibitem{papineni2002bleu}
Papineni, K., Roukos, S., Ward, T., Zhu, W.J.: Bleu: a method for automatic evaluation of machine translation. In: Proceedings of the 40th annual meeting of the Association for Computational Linguistics. pp. 311--318 (2002)

\bibitem{pennington2014glove}
Pennington, J., Socher, R., Manning, C.D.: Glove: Global vectors for word representation. In: Proceedings of the 2014 conference on empirical methods in natural language processing (EMNLP). pp. 1532--1543 (2014)

\bibitem{raffel2020exploring}
Raffel, C., Shazeer, N., Roberts, A., Lee, K., Narang, S., Matena, M., Zhou, Y., Li, W., Liu, P.J., et~al.: Exploring the limits of transfer learning with a unified text-to-text transformer. J. Mach. Learn. Res.  \textbf{21}(140),  1--67 (2020)

\bibitem{ranganathan2016multimodal}
Ranganathan, H., Chakraborty, S., Panchanathan, S.: Multimodal emotion recognition using deep learning architectures. In: 2016 IEEE Winter Conference on Applications of Computer Vision (WACV). pp.~1--9. IEEE (2016)

\bibitem{rashkin2019towards}
Rashkin, H., Smith, E.M., Li, M., Boureau, Y.L.: Towards empathetic open-domain conversation models: A new benchmark and dataset. In: Proceedings of the 57th Annual Meeting of the Association for Computational Linguistics. pp. 5370--5381 (2019)

\bibitem{russell1980circumplex}
Russell, J.A.: A circumplex model of affect. Journal of personality and social psychology  \textbf{39}(6), ~1161 (1980)

\bibitem{russell2019human}
Russell, S.: Human compatible: Artificial intelligence and the problem of control. Penguin (2019)

\bibitem{sammani2022nlx}
Sammani, F., Mukherjee, T., Deligiannis, N.: Nlx-gpt: A model for natural language explanations in vision and vision-language tasks. In: Proceedings of the IEEE/CVF Conference on Computer Vision and Pattern Recognition. pp. 8322--8332 (2022)

\bibitem{sap2019social}
Sap, M., Rashkin, H., Chen, D., Le~Bras, R., Choi, Y.: Social iqa: Commonsense reasoning about social interactions. In: Proceedings of the 2019 Conference on Empirical Methods in Natural Language Processing and the 9th International Joint Conference on Natural Language Processing (EMNLP-IJCNLP). pp. 4463--4473 (2019)

\bibitem{seo2017visual}
Seo, P.H., Lehrmann, A., Han, B., Sigal, L.: Visual reference resolution using attention memory for visual dialog. Advances in neural information processing systems  \textbf{30} (2017)

\bibitem{sharma2018conceptual}
Sharma, P., Ding, N., Goodman, S., Soricut, R.: Conceptual captions: A cleaned, hypernymed, image alt-text dataset for automatic image captioning. In: Proceedings of ACL (2018)

\bibitem{simonyan2014very}
Simonyan, K., Zisserman, A.: Very deep convolutional networks for large-scale image recognition. arXiv preprint arXiv:1409.1556  (2014)

\bibitem{strapparava2007semeval}
Strapparava, C., Mihalcea, R.: Semeval-2007 task 14: Affective text. In: Proceedings of the Fourth International Workshop on Semantic Evaluations (SemEval-2007). pp. 70--74 (2007)

\bibitem{touvron2023llama}
Touvron, H., Martin, L., Stone, K., Albert, P., Almahairi, A., Babaei, Y., Bashlykov, N., Batra, S., Bhargava, P., Bhosale, S., et~al.: Llama 2: Open foundation and fine-tuned chat models. arXiv preprint arXiv:2307.09288  (2023)

\bibitem{mephisto}
Urbanek, J., Ringshia, P.: Mephisto: A framework for portable, reproducible, and iterative crowdsourcing (2023). \doi{10.48550/ARXIV.2301.05154}, \url{https://arxiv.org/abs/2301.05154}

\bibitem{Vries_2017_CVPR}
de~Vries, H., Strub, F., Chandar, S., Pietquin, O., Larochelle, H., Courville, A.: Guesswhat?! visual object discovery through multi-modal dialogue. In: Proceedings of the IEEE Conference on Computer Vision and Pattern Recognition (CVPR) (July 2017)

\bibitem{Weng_2023_ICCV}
Weng, S., Zhang, P., Chang, Z., Wang, X., Li, S., Shi, B.: Affective image filter: Reflecting emotions from text to images. In: Proceedings of the IEEE/CVF International Conference on Computer Vision (ICCV). pp. 10810--10819 (October 2023)

\bibitem{yanulevskaya2008emotional}
Yanulevskaya, V., van Gemert, J.C., Roth, K., Herbold, A.K., Sebe, N., Geusebroek, J.M.: Emotional valence categorization using holistic image features. In: 2008 15th IEEE international conference on Image Processing. pp. 101--104. IEEE (2008)

\bibitem{you2016building}
You, Q., Luo, J., Jin, H., Yang, J.: Building a large scale dataset for image emotion recognition: The fine print and the benchmark. In: Proceedings of the AAAI conference on artificial intelligence. vol.~30 (2016)

\bibitem{young2014image}
Young, P., Lai, A., Hodosh, M., Hockenmaier, J.: From image descriptions to visual denotations: New similarity metrics for semantic inference over event descriptions. Transactions of the Association for Computational Linguistics  \textbf{2},  67--78 (2014)

\bibitem{yuan2021bartscore}
Yuan, W., Neubig, G., Liu, P.: Bartscore: Evaluating generated text as text generation. Advances in Neural Information Processing Systems  \textbf{34},  27263--27277 (2021)

\bibitem{zhang2019bertscore}
Zhang, T., Kishore, V., Wu, F., Weinberger, K.Q., Artzi, Y.: Bertscore: Evaluating text generation with bert. In: International Conference on Learning Representations (2019)

\bibitem{zhou2023sotopia}
Zhou, X., Zhu, H., Mathur, L., Zhang, R., Yu, H., Qi, Z., Morency, L.P., Bisk, Y., Fried, D., Neubig, G., et~al.: Sotopia: Interactive evaluation for social intelligence in language agents. arXiv preprint arXiv:2310.11667  (2023)

\bibitem{zhu2023chatgpt}
Zhu, D., Chen, J., Haydarov, K., Shen, X., Zhang, W., Elhoseiny, M.: Chatgpt asks, blip-2 answers: Automatic questioning towards enriched visual descriptions. arXiv preprint arXiv:2303.06594  (2023)

\end{thebibliography}
\end{document}